\documentclass{article}

\usepackage[preprint]{neurips_2024}

\usepackage[utf8]{inputenc} 
\usepackage[T1]{fontenc}    
\usepackage{hyperref}       
\usepackage{url}            
\usepackage{booktabs}       
\usepackage{amsfonts}       
\usepackage{nicefrac}       
\usepackage{microtype}      
\usepackage{xcolor}         

\usepackage{wrapfig}

\usepackage{microtype}
\usepackage{graphicx}
\usepackage{subfigure}
\usepackage{booktabs} 
\usepackage[normalem]{ulem}
\usepackage{adjustbox}
\usepackage{amsmath}
\usepackage{amssymb}
\usepackage{mathtools}
\usepackage{array}
\usepackage{enumitem}
\usepackage{subfigure}
\usepackage{colortbl}
\usepackage{soul}
\usepackage{multirow}
\usepackage{makecell}
\definecolor{mypink}{rgb}{.99,.90,.92}
\definecolor{magicmint}{rgb}{0.67, 0.94, 0.82}

\usepackage{listings}
\usepackage{pythonhighlight}

\usepackage{titletoc}
\usepackage{letltxmacro}
\usepackage{etoolbox}
\usepackage{lipsum}

\definecolor{mypink}{rgb}{.99,.91,.95}
\definecolor{mygrey}{rgb}{.9,.9,.9}

\usepackage{soul,color,xcolor}
\usepackage[capitalize,noabbrev]{cleveref}
\colorlet{soulblue}{blue!20}
\colorlet{yw}{orange!50}
\definecolor{mypurple}{rgb}{.80, .70, .84}
\definecolor{myblue}{rgb}{.86, .91, .95}

\title{ RespLLM: Unifying Audio and Text with Multimodal LLMs for Generalized Respiratory Health Prediction}

\author{%
  Yuwei Zhang$^1$, 
  \And Tong Xia$^1$, 
 \And Aaqib Saeed$^2$, 
  \And Cecilia Mascolo$^1$ \\
\And $^1$ University of Cambridge, UK$^2$ Eindhoven University of Technology, Netherlands\\
  \texttt{yz798@cam.ac.uk} \\
}

\begin{document}

\maketitle

\vspace{-15pt}
\begin{abstract}
  The high incidence and mortality rates associated with respiratory diseases underscores the importance of early screening. Machine learning models can automate clinical consultations and auscultation, offering vital support in this area. However, the data involved, spanning demographics, medical history, symptoms, and respiratory audio, are heterogeneous and complex. Existing approaches are insufficient and lack generalizability, as they typically rely on limited training data, basic fusion techniques, and task-specific models. In this paper, we propose RespLLM, a novel multimodal large language model (LLM) framework that unifies text and audio representations for respiratory health prediction. RespLLM leverages the extensive prior knowledge of pretrained LLMs and enables effective audio-text fusion through cross-modal attentions. Instruction tuning is employed to integrate diverse data from multiple sources, ensuring generalizability and versatility of the model. Experiments on five real-world datasets demonstrate that RespLLM outperforms leading baselines by an average of 4.6\% on trained tasks, 7.9\% on unseen datasets, and facilitates zero-shot predictions for new tasks. Our work lays the foundation for multimodal models that can \textit{perceive}, \textit{listen to}, and \textit{understand} heterogeneous data,  paving the way for scalable respiratory health diagnosis.
\end{abstract}

\section{Introduction}
\label{sec:intro}


\begin{wrapfigure}{r}{7cm}
\vspace{-25pt}
\centering
{\includegraphics[width=0.9\linewidth]{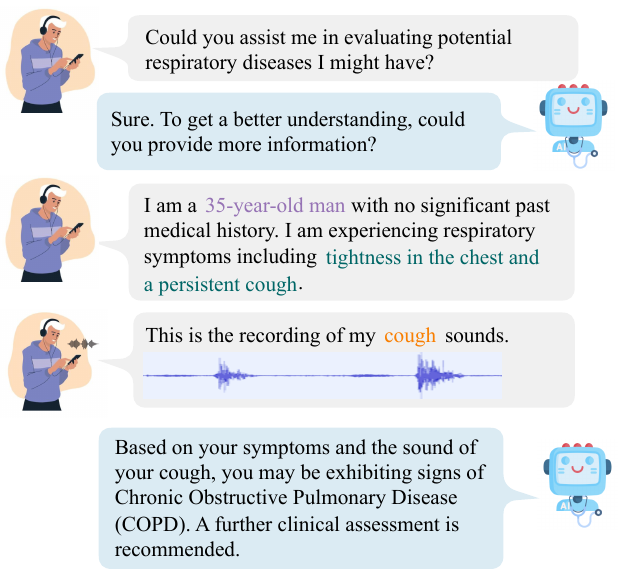}}
\caption{Automated consultation and auscultation for respiratory health screening.} 
\label{fig:interaction}
\vspace{-20pt}
\end{wrapfigure}

Respiratory diseases are the third leading cause of death worldwide, highlighting the critical need for early and accessible respiratory health screening~\citep{labaki2020chronic}. Clinical assessment of such diseases typically begins with gathering personal information (\textit{consultation}), including demographics, medical history, symptoms, and other relevant details (hereafter collectively referred to as DMS). In addition, clinicians listen to respiratory sounds (\textit{auscultation}) as a non-invasive method of screening, before proceeding to more invasive and costly examinations~\citep{Reyes2024}. Consequently, automating both the consultation and auscultation processes using machine learning (ML), as illustrated in \cref{fig:interaction}, can significantly enhance early screening by increasing efficiency, accessibility, and affordability.

\begin{wrapfigure}{r}{7.5cm}
  {\includegraphics[width=0.95\linewidth]{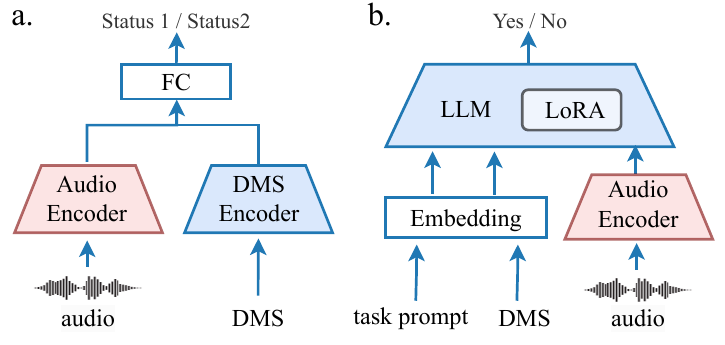}}
    \caption{\textbf{Multimodal models for respiratory health prediction.} (a) Existing concatenation-based fusion method. (b) Our LLM-based fusion method. 
  }
  \label{fig:1}
\end{wrapfigure}

\vspace{5pt}
Considering that the DMS and audio data are different modalities, presenting heterogeneous information, multimodal ML approaches that can effectively integrate them are needed.
Early efforts have been made in this direction~\citep{xia2023cross,kim2024bts,han2021exploring}; nevertheless, limitations hinder their application in real-world diagnostic scenarios. 
First, \textit{these models are typically small-scale and trained on limited data}, restricting their ability to effectively learn from high-dimensional audio signals and unstructured DMS data.
Second, \textit{the fusion of DMS and audio remains inadequate, which may reduce model performance}. They commonly concatenate audio representations with DMS representations, encoded either into categorical vectors using a pre-defined mapping~\citep{han2021exploring} or into word embeddings~\citep{kim2024bts}. Such concatenation overlooks the differences in their embedding spaces and interrelationship between the two types of data.

\vspace{5pt}
More importantly, \textit{existing models are task- and dataset-specific, which hinders their ability to generalize}. Traditional machine learning models rely on the IID  (Independent and Identically Distributed) assumption, and when the data distribution shifts, their performance tends to degrade. However, respiratory health data for model training is often limited~\citep{kim2024bts}, and in real-world deployments, data ranging from DMS to audio, as well as the respiratory status included, can differ significantly from the training data. For example, a model trained to predict asthma may be required to predict COVID-19 status at the inference stage. Highly generalized models capable of handling these changes are necessary but currently lacking.

\vspace{5pt}
To overcome these limitations and progress towards the envisioned applications depicted in \cref{fig:interaction}, this paper puts forward a unique approach that harnesses the power of pre-trained LLMs to simultaneously interpret DMS and audio for respiratory health screening. The high-level concept of the proposed method is illustrated in \cref{fig:1}b. 
Unlike existing methods, which are constrained by limited data and model scale,  our approach leverages LLMs extensively trained on large corpora, including medical materials~\citep{goel2023llms}, to extend model capacity beyond the available respiratory training data.
For effective multimodal fusion, we generate sequences of audio representations from a pre-trained encoder and combine them with DMS text embeddings as a unified input to the LLM. This enables coherent integration of the two modalities through multi-layer and multi-head attention mechanisms. 
To enhance the model generalizability, we curate multiple data sources for training and create instructions applicable to a variety of tasks that combine DMS and audio. This approach equips the model with zero-shot inference capabilities for new datasets and unseen tasks.

\vspace{5pt}
Our contributions can be summarized as follows:
\begin{enumerate}
\item 
To the best of our knowledge, this work presents, for the first time, the use of LLMs to jointly model DMS and audio data for respiratory health screening. The proposed multimodal LLM, RespLLM, can comprehensively \textit{perceive}, \textit{listen to}, \textit{understand} heterogeneous inputs and then \textit{diagnose} respiratory health.
\item We curate a large instruction-tuning set combining task prompts, DMS, and audio  to optimize the proposed model. This approach ensures the model remains versatile (one model for multiple tasks) and generalized (performing well on new datasets or tasks).
\item We conduct extensive experiments on multiple open datasets. Results demonstrate the superiority of our model over existing methods, showing notable improvement in both trained and unseen tasks, 
along with the robustness of our approach in integrating different LLM models. 
\end{enumerate}

\section{Related Work}


\subsection{ML for Respiratory Health}
In clinical practice, respiratory health is assessed through various clinical examinations such as spirometry, auscultation, chest X-rays, plethysmography, and computed tomography scans~\citep{Reyes2024}. 
Auscultation, combined with personal DMS information, is
among 
the most comfortable and affordable approaches. Using an electronic stethoscope or a microphone, respiratory sounds, such as coughing and breathing, produced by airflow in the respiratory system can be easily recorded. These recordings contain valuable physiological information related to breathing difficulties, reduced oxygen saturation, and other conditions~\citep{xia2022exploring}. Therefore, modeling respiratory audio and DMS data holds significant potential for ubiquitous respiratory health monitoring.

Traditionally, audio signal processing techniques were used to extract acoustic features that help distinguish between different respiratory conditions~\citep{ma2022determining,islam2018multichannel}. Recently, deep learning (DL) has significantly advanced acoustic modeling by automatically capturing complex relationships from raw audio data or spectrograms. This advancement has led to high-performing applications, from detecting abnormal lung sounds to diagnosing conditions such as the flu and pulmonary diseases~\citep{gairola2021respirenet,fraiwan2022recognition,srivastava2021deep}. When combined with additional information like DMS, DL-driven respiratory health prediction models demonstrate further performance improvements~\citep{han2021exploring,xia2023cross,kim2024bts,moummad2023pretraining}.

\begin{wrapfigure}{r}{7cm}
\vspace{-10pt}
  {\includegraphics[width=0.9\linewidth]{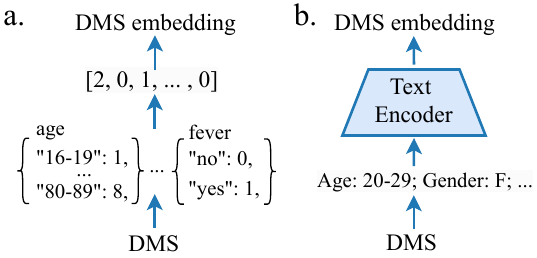}}
  \caption{\textbf{Existing DMS encoding methods.} (a) Pre-defined mapping. (b) Text embedding.}
  \label{fig:dms}
\end{wrapfigure}

However, current methods to represent and fuse DMS and audio in the field of respiratory health remain simple and may fail to capture all the relevant information. DMS is typically encoded either by mapping variables into a uniform vector using a predefined dictionary  (\cref{fig:dms}a)~\citep{han2021exploring,xia2023cross} or by extracting text embeddings from the unstructured data (\cref{fig:dms}b)~\citep{kim2024bts,moummad2023pretraining}. This representation is then concatenated with audio from a deep learning encoder, ignoring the differences and complex relationship between the two, limiting the potential of DL for health prediction. 
In the related field of chest X-ray modeling, more advanced multimodal techniques such as LSTM-based fusion~\citep{hayat2022medfuse}, cross-modal attention~\citep{wang2018tienet}, and multimodal pre-training~\citep{moon2022multi} have been explored. 
In this paper we explore how similar approaches could be beneficial to audio and DMS. 

\subsection{LLMs for Health}
Recently-emerged LLMs have demonstrated remarkable capabilities in various health diagnostic applications~\citep{singhal2023large,  lievin2024can}. This is primarily due to their pretraining on enormous and diverse datasets, including medical literature, clinical guidelines, research papers, and general knowledge~\citep{goel2023llms}. Such pretraining enables LLMs to understand medical terminology, concepts, and associations relevant to health diagnostics.

There is also a growing trend in extending LLMs, which are inherently language models, to handle multimodal data in a unified manner~\citep{wu2023next, qiu2023automated}. This capability is typically achieved by combining prompts, modality-specific encoders, and LLMs within a single framework~\citep{moor2023med, yu2023zero,liuzero}. For example, Liu \textit{et al.}~\cite{liuzero} leveraged LLMs to interpret electrocardiography signals and perform zero-shot diagnosis. 
To further enhance generalizability, instruction tuning has emerged as a promising approach for adapting LLMs to various tasks and domains~\citep{aw2023instruction}. 
In this work, we make the first effort to leverage recent advancements in multimodal LLMs and curate an instruction-tuning dataset using diverse sources for generalized respiratory health prediction.
\begin{figure*}[t]
  {\includegraphics[width=0.99\linewidth]{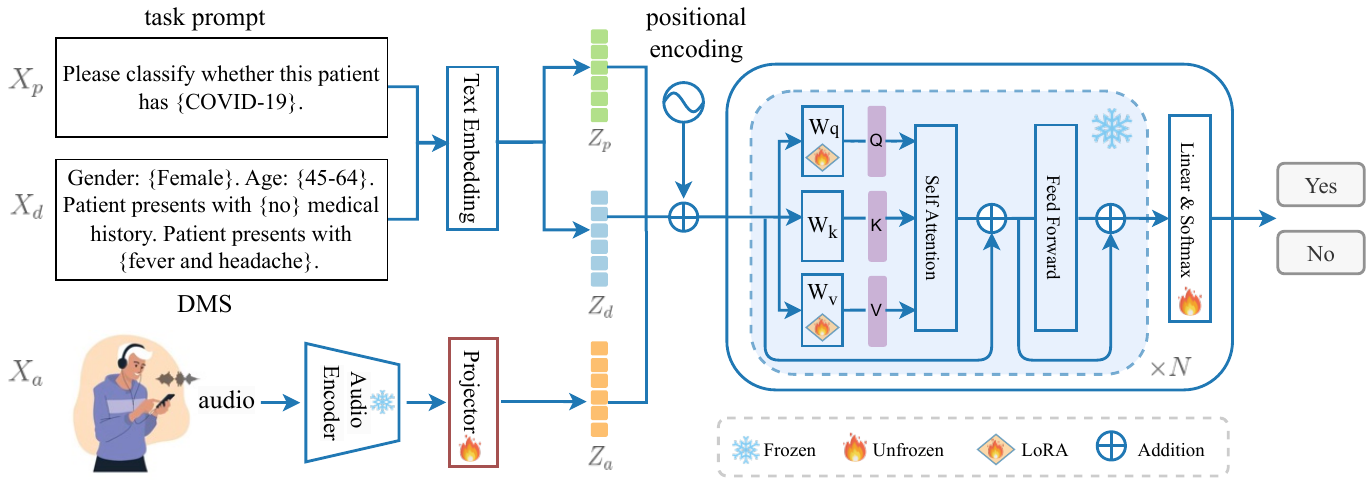}}
  \caption{\textbf{The model architecture of RespLLM.} Text embeddings from task prompts and personal DMS, along with audio embeddings from respiratory sounds, are sequentialized as input for the LLM consisting of multiple transformer blocks.}
  \label{fig:model}
\end{figure*}

\section{Methodology}

\cref{fig:1}b illustrates our proposed framework, a multimodal LLM that can model DMS and respiratory audio simultaneously. In this section, we begin by elaborating on the model architecture design. Then we delve into how we curate the instruction tuning
dataset to train this model. 


\subsection{Model Architecture}
Our overall model architecture is shown in \cref{fig:model}.  Given the DMS $X_{d}$ and the respiratory audio signal $X_{a}$, our goal is to provide a screening result/recommendation
in response to the question in the prompt $X_{p}$. To achieve this,  our model mainly consists of three modules: a text embedder that maps $X_{p}$ and $X_{d}$ into text token embeddings, an audio encoder with a projector to map $X_{a}$ into audio embeddings, and an LLM to fuse all the given information for respiratory health screening. These modules are specified as follows. 



\paragraph{Text embedding.} The text embedding module will first split the given prompt $X_{p}$ and DMS $X_{d}$ into sequence of tokens using its tokenizer, and then map the words into a sequence of word embeddings, denoted by $Z_p\in \mathbb{R}^{L_p \times S}$ and $Z_d\in \mathbb{R}^{L_d \times S}$, where $L_p$ and $L_d$ are the lengths of the text and $S$ is the dimension of the word embeddings.  For consistency, we use the same tokenizer and word embeddings from the LLM that is used in the later stage. In this sense, $S$ is also the dimension of the hidden state in the transformer blocks of the used LLM.

\paragraph{Audio Encoder with Projector.}
Given the high dimensionality and complexity of the audio data, we adapt a pre-trained audio encoder to obtain audio embeddings for $X_a$~\citep{zhang2024towards}. Each audio sample is first transformed into a spectrogram, which is then divided into small patches of equal size to derive embeddings. 
We feed the resulting sequence of $L_a$ embeddings into the LLM, denoted by $z_a \in \mathbb{R}^{L_a \times A}$, where $A$ is the dimension of the original audio embeddings.
As the LLM  has a different hidden embedding space of dimension $S$, we need to efficiently align the audio embeddings with word embeddings. Following insights from previous work~\citep{ma2024embarrassingly}, we use a simplistic linear layer as the projector $\cal{P}(\cdot)$. Then, we have the final audio embeddings $Z_a = {\cal{P}}{(z_{a})}$, where $Z_a \in \mathbb{R}^{L_a \times S}$.




\paragraph{LLM and LoRA.} 

For the three distinct embedding $Z_p$, $Z_d$, and $Z_a$, which correspond to task prompt, DMS, and audio information respectively, we first combine them into a longer sequence of embeddings. After this, we add positional embeddings to the resulting sequence, producing the final embedding $Z\in \mathbb{R}^{L \times S}, L = L_p + L_d + L_a$. Note that we use the same positional embedding approach as that employed by the chosen LLM model. This combined embedding $Z$ is then fed into the LLM for further processing. 

Since the LLM consists of multiple transformer blocks as shown by the blue shaded box in \cref{fig:model}, each containing several self-attention operations parameterized by $W_q, W_k$ and $W_v$, the three types of information are deeply fused. The final transformer block outputs a sequence of hidden states with a length of $L$, which are then flattened across the temporal dimension to generate a single vector representation. This vector is then passed through a linear layer with a \textit{Softmax} function to produce the final output, yielding binary health predictions. To mitigate the risk of hallucinations in the original LLM output, we replace the original linear layer with a randomly initialized one containing two output nodes, representing the answer, either `Yes' or `No', to the question in the task prompt.

To balance between preserving the LLM's prior knowledge from large-scale pretraining and adapting it to respiratory health prediction tasks, we choose to update only part of the pretrained parameters. LoRA (Low-Rank Adaptation)~\citep{hu2021lora} is a parameter-efficient fine-tuning method that reduces the computational cost of updating large models. As shown in \cref{fig:model}, we apply LoRA to the key ($W_k$) and query ($W_q$) mapping modules in the transformer blocks of the LLM, while keeping the rest of the parameters frozen. 




\subsection{Model Training}
\begin{figure*}[t]
  {\includegraphics[width=0.99\linewidth]{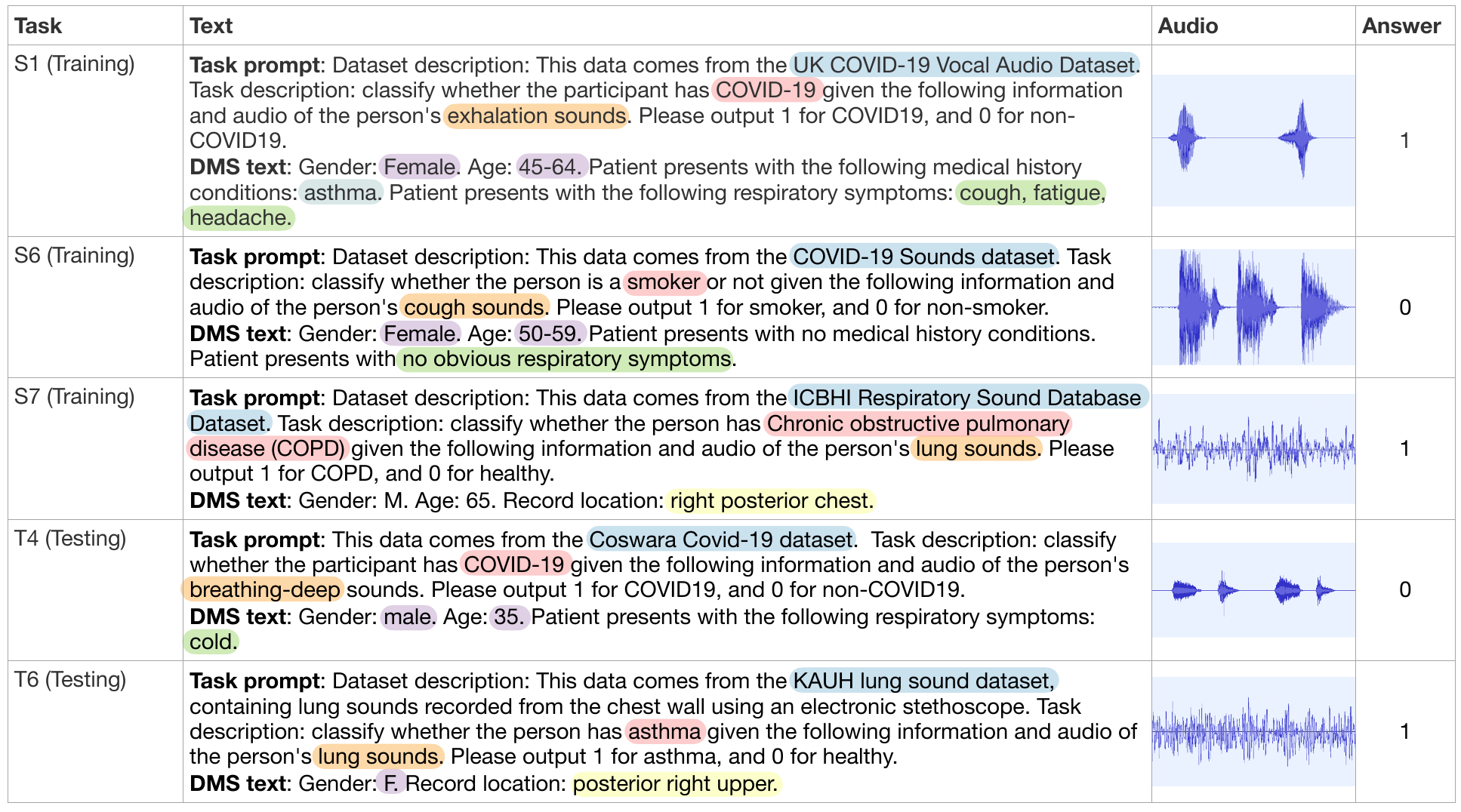}}
  \caption{\textbf{Examples of instructions used in our work.} The variables that differ across samples and datasets are highlighted. For any missing data in a field, the corresponding description is omitted.}
  \label{fig:prompt}
\end{figure*}

\paragraph{Data Curation.}
To increase the generality of our method, we propose to combine multiple data resources for training. Those data can differ in the audio modalities, DMS formats and the category of respiratory conditions. To unify them for model training, we design contextualized instructions containing task prompts, the description of DMS and the corresponding audio information. The templates of $X_{p}$ and $X_{d}$ are formulated as described below, with examples provided in \cref{fig:prompt}. 

\vspace{5pt}
\noindent I. The task prompt $X_{p}$ is a diagnostic  query with respect to the condition that can be predicted from the given audio and DMS. It is formulated as: 
\begin{quote}
\small
\vspace{-3pt}
 ``\textit{\textbf{Dataset description}: This data comes from the \{D\}. \textbf{Task description}: classify whether the participant has \{C\} given the following information and audio of the person's \{T\} sounds. Please output 1 for \{C1\}, and 0 for \{C2\}.
 }"
 \vspace{-3pt}
\end{quote}
Here, D distinguishes the data resource, T presents the sound type, and C denotes the condition to be predicted from C1 and C2 restricts the output space. 

\vspace{5pt}
\noindent II. For the text input of DMS $X_{d}$, we use the following template: 
\begin{quote}
\small
\vspace{-3pt}
``\textit{\textbf{Gender}: \{G\}. \textbf{Age}: \{A\}. Patient presents with \{M\} \textbf{medical history} conditions. Patient presents with the following \textbf{respiratory symptoms}: \{S\}. Recorded location: \{L\}.} "
\vspace{-3pt}
\end{quote}
Here, G denotes the gender, A represents age, M specifies medical history, and S is the list of symptoms. L represents the location where the audio was recorded for lung sounds.
For any missing or non-applicable data field, the corresponding description is omitted.

\begin{table}[t]
\centering
  \caption{\textbf{Summary of \textit{source} and \textit{target} datasets and tasks used in this study.} The five datasets are UK COVID, COVID-19 Sounds, ICBHI, Coswara, and KAUH. For task IDs, S1$-$S7 refer to the source tasks, and T1$–$T6 refer to the target tasks. In audio types, `s' is short for shallow, `h' for heavy, and `d' for deep.}
  \begin{tabular}{clllll}
  \toprule
  \bfseries Data& \bfseries ID  & \bfseries Label  & \bfseries Audio Type & \bfseries  \#Train/Test\\
  
  \midrule
1 & S1      & Covid           & Exhalation    & 1500/1000     \\
1 & S2      & Covid           & Cough         & 1500/1000     \\
2 & S3      & Covid           & Breath        & 1162/324      \\
2 & S4      & Covid           & Cough         & 1162/324      \\
2 & S5      & Smoker          & Breath        & 2570/1419     \\
2 & S6      & Smoker          & Cough         & 2570/1419     \\
3 & S7      & COPD            & Lung sounds   & 462/366       \\
\midrule
4 & T1      & Covid           & Cough-s     & -/100      \\
4 & T2      & Covid           & Cough-h     & -/100      \\
4 & T3      & Covid           & Breath-s & -/40     \\
4 & T4      & Covid           & Breath-d & -/40     \\
5 & T5      & COPD            & Lung sounds   & -/38         \\ 
5 & T6      & Asthma          & Lung sounds   & -/116   \\
  \bottomrule
  \end{tabular}
  \label{tab:task}
\end{table}

\paragraph{Instruction Tuning.} Since various data resources have been unified into instructions, we can now shuffle these instructions from multiple sources to create batches for model training. To make the most of the pre-trained knowledge in the audio encoder and the LLM, we will only train the projector, the LoRA parameters, and the final fully connected layer for the LLM in our model, as shown in \cref{fig:model}. For the objective function, we use the cross-entropy loss, comparing the output of the LLM with the actual answer to the diagnostic question in the prompt. 





\paragraph{Zero-shot Prediction.}
As mentioned earlier, since the diagnostic task and personal DMS are formulated in text, our model can easily extend to new data and unseen respiratory conditions. This allows for zero-shot inference without requiring any parameter changes when deploying to a new domain.

\section{Experiments}

In this section, we conduct extensive experiments with real-world data to answer the following questions:

\begin{itemize}[topsep=-1pt, itemsep=0pt, leftmargin=15pt]
    \item  \textbf{RQ1:} How does our model perform compared to the state-of-the-art baselines for respiratory health prediction?
    \item  \textbf{RQ2:} How well does our model generalize to new data and unseen tasks?
    \item  \textbf{RQ3:} How do the model design and the choice of LLMs impact the performance of our method?
    
\end{itemize}

\subsection{Datasets and Tasks}
 
We use five open respiratory audio datasets for our experiments, featuring recordings of coughing, breathing, and lung sounds related to respiratory health statuses like smoking, COVID-19, and other respiratory diseases. These datasets also contain rich DMS information including age, gender, medical histories, symptoms, and recording locations for lung sounds.
Using these datasets, we define 13 respiratory health tasks, as shown in \cref{tab:task}. Among these, only the source tasks are used for model training, while the others are reserved for testing. Examples of the instructions we generated by combining task prompts, DMS, and audio recordings are illustrated in \cref{fig:prompt}.

\begin{table*}[t]
\centering
\footnotesize
\caption{\textbf{Performance when training and testing on the same data sources.} Baselines are task-specific (trained and tested on each single task), while RespLLM is trained collectively with all tasks. Best results are bold and the second best are underlined. }
\vspace{5pt}
\begin{adjustbox}{max width=\linewidth}
\begin{tabular}{llcccccccc}
  \toprule

& \bfseries Task                   & \bfseries S1 $\rightarrow$ S1    &  \bfseries S2$\rightarrow$S2    & \bfseries S3$\rightarrow$S3           & \bfseries S4$\rightarrow$S4         & \bfseries S5$\rightarrow$S5     & \bfseries S6$\rightarrow$S6     & \bfseries S7$\rightarrow$S7     & \bfseries \textit{Avg.}    \\
  \midrule
Single-modal & Audio              & 0.6025 & 0.6729 & 0.5828        & 0.6260     & 0.5517 & 0.6247 & 0.9575 & 0.6597 \\
 & DMS - Hard   & 0.7626 & 0.7521 & 0.6427        & 0.6427     & 0.5485 & 0.5485 & 0.8341 & 0.6759 \\
 & DMS - Soft & \underline{0.9126} & 0.8900 & \underline{0.7406}    & \underline{0.7406}   & 0.5594 & 0.5594 & 0.9938 & 0.7709 \\
\midrule
Multimodal &  Fusion - Hard           & 0.5936 & 0.6905 & 0.6171        & 0.6747     & \underline{0.5714} & 0.6250 & 0.9845 & 0.6795 \\
 & Fusion - Soft & 0.8668          & \underline{0.8954}          & 0.6997          & 0.7390          & 0.5692          & \textbf{0.6336} & \underline{0.9981}          & \underline{0.7717} \\
 
   & RespLLM (Ours) & \textbf{0.9244}         & \textbf{0.9002}         & \textbf{0.7958}         & \textbf{0.7840}         & \textbf{0.6189 }        & \underline{0.6274}          & \textbf{1.0000}        & \textbf{0.8072}  \\
  \bottomrule
  \end{tabular}
  \end{adjustbox}
  \label{tab:supervised}
\end{table*}

\begin{table*}[t]
\footnotesize
\caption{\textbf{Performance of zero-shot prediction on new datasets.} For baselines, Sx presents the models used for testing, i.e., S2\&4$\rightarrow$T1, S2\&4$\rightarrow$T2, S1\&3$\rightarrow$T3, S1\&3$\rightarrow$T4, S7$\rightarrow$T5. The average performance is reported when multiple models can be transferred. For RespLLM, Sx refers to our trained model with all source task data.}
\vspace{5pt}
\begin{adjustbox}{max width=\linewidth}
    
    \begin{tabular}{llcccccc|cc}
  \toprule

& \bfseries Task                   & \bfseries Sx$\rightarrow$T1    &  \bfseries Sx$\rightarrow$T2    & \bfseries Sx$\rightarrow$T3           & \bfseries Sx$\rightarrow$T4         & \bfseries Sx$\rightarrow$T5     & \bfseries \textit{Avg.}         & \bfseries Sx$\rightarrow$T6    \\
  \midrule

  Single-modal & Audio          & 0.6076 & 0.4940 & 0.5963 & 0.4875 & 0.5823 & 0.5535 & -      \\
             & DMS - Hard & 0.4956 & 0.4956 & \underline{0.6312} & 0.6312 & 0.5375 & 0.5582 & -      \\
             & DMS - Soft & 0.5834 & 0.5834 & 0.5525 & 0.5525 & 0.5312 & 0.5606 & -      \\
\midrule
Multimodal   & Fusion - Hard  & 0.5528 & 0.5276 & 0.6288 & 0.5550 & 0.5708 & 0.5670 & -      \\
             & Fusion - Soft  & \underline{0.6190} & \underline{0.5928} & 0.6288 & \underline{0.6400} & 0.5542 & \underline{0.6070} & -      \\
             & RespLLM (Ours) & \textbf{0.6424} & \textbf{0.6284} & \textbf{0.6525} & \textbf{0.6750} & \textbf{0.6750} & \textbf{0.6547} & 0.5865 \\
  
  \bottomrule
  \end{tabular}
  \end{adjustbox}
  \label{tab:zeroshot}
\end{table*}

\subsection{Experimental Setup}
For comparison, we implement both single-modal and multimodal baselines. 
Regarding single-modal methods, we compare to \textbf{\textit{Audio}}, which fine-tunes the pre-trained audio encoder alongside a linear classifier for respiratory condition prediction~\citep{xia2021covid}.
For DMS-only methods, we consider to use the hard encoding in \cref{fig:dms}a
and soft text embedding in \cref{fig:dms}b to fit a linear model, namely \textbf{\textit{DMS-hard}} and \textbf{\textit{DMS-soft}}, respectively. Based on these two methods for DMS, we compare to the multimodal method as illustrated in \cref{fig:1}a, and name them  \textbf{\textit{Fusion-hard}}~\citep{han2021exploring} and \textbf{\textit{Fusion-soft}}~\citep{kim2024bts}, as our multimodal baselines.

The audio encoder used in both the baselines and our method is the pre-trained OPERA-CT model~\citep{zhang2024towards}, a hierarchical token-semantic audio transformer. It processes an 8-second audio input (padded or cropped) into a spectrogram of size $256 \times 64$ and output embeddings of 64 patches, each with a dimension of $768$. 
The LLM model that we modify is OpenBioLLM-8B\footnote{\url{https://huggingface.co/aaditya/Llama3-OpenBioLLM-8B}} which is an open-source LLM designed for the biomedical domain.
The instruction tuning is completed
on a single A-100 GPU. For all tasks, we use AUROC as the metric to report the health condition prediction performance.

\subsection{Results}

\paragraph{Health Prediction Performance~(RQ1).}

To answer RQ1, we first examine the performance of our model and the baselines when testing on training datasets (held-out testing set). Since the baselines are task-specific by design, they are trained and tested on the same task, whereas our model utilizes all data resources, resulting in a single RespLLM capable of performing well on multiple tasks. The results are summarized in  \cref{tab:supervised}.  Among the seven evaluated tasks, our model outperforms the state-of-the-art baselines on six tasks, with the average AUROC across all seven tasks surpassing the best baseline by 4.6\% (0.8072 vs. 0.7717). It can also be observed that the fusion baselines compared cannot consistently outperform their single-modal counterparts, and their average AUROCs are very close. This suggests that the fusion methods are insufficient. In contrast, our model demonstrates superiority by effectively fusing DMS and audio information via the LLM for respiratory health prediction.

\paragraph{Generalizability~(RQ2).} To demonstrate our model's generalizability and address RQ2, we evaluate its performance not only on in-distribution data but also on new, unseen datasets and tasks.
Specifically, we train the models on source task data and test them on target tasks. As shown in \cref{tab:task} and \cref{fig:prompt}, both the types of sounds and the information from DMS vary between source and target tasks. Our model can be directly tested, while for the baselines, we report cross-task transfer performance. Since no fine-tuning is applied, this constitutes zero-shot prediction, with the results summarized in \cref{tab:zeroshot}. 

Zero-shot transfer prediction shows a degraded performance compared to \cref{tab:supervised}, due to changes in data sources, audio types, and DMS information. Despite this challenge, our model consistently outperforms all compared baselines, with the average AUROC surpassing the best baseline by 7.9\% (0.6547 vs. 0.6070). This demonstrates the stronger generalizability of our method over the baselines. 
Notably, in T6, where asthma is a new class not included in the training data, none of the baselines can predict this condition (e.g., a model trained to distinguish COVID/non-COVID in S1 cannot differentiate asthma from healthy cases). In contrast, our model achieves an AUROC of 0.5865, comparable to the baselines' average performance on T1-5.
This capability largely stems from our instruction-tuning approach, which effectively retrieves relevant knowledge from the pretrained LLM for zero-shot generalization.

\paragraph{Effect of Training and Model Design (RQ3). } 

To further validate the superiority of our framework with cross-data training, we perform several ablation studies. We combine S1-7 into a multi-label task and use all data to train the multimodal baselines for direct comparison of different fusion methods: concatenation fusion as used in the baseline, add-on fusion from~\citep{blandfort2019fusion}, and cross-attention fusion from~\citep{wang2022cross}. The results for normal testing on source tasks and zero-shot prediction on target tasks are shown in \cref{tab:ablation:fusion:supervised} and \cref{tab:ablation:fusion:zeroshot}. Concatenation outperforms addition, as the audio and text embeddings are in very different spaces, and simply adding them may confuse the model. 
Concatenation also outperforms cross-attention fusion, likely because attention introduces additional parameters to train, which increases the data demand. Our model outperforms all these ablations due to the use of more complex architectures with pretrained parameters and knowledge.

\begin{table*}[hbtp]
\footnotesize
\caption{Performance of different fusion methods in our framework when testing on source datasets.}
\vspace{3pt}
\begin{tabular}{lcccccccc}
  \toprule

 Task                   & \bfseries S1    &  \bfseries S2    & \bfseries S3           & \bfseries S4         & \bfseries S5     & \bfseries S6      & \bfseries S7 & \bfseries Avg.    \\
  \midrule
\textbf{Fusion - Soft}      & 0.9065      & 0.8927      & 0.7436      & 0.7396      & 0.5884      & 0.5833      & 0.9543       & 0.7726 \\
\textbf{Fusion - Add}       & 0.7525      & 0.7828      & 0.7289      & 0.7223      & 0.5653      & 0.5930      & 0.6941       & 0.6913 \\
\textbf{Fusion - CrossAttn} & 0.8131      & 0.8369      & 0.7870      & 0.7805      & 0.5754      & 0.5872      & 0.7942       & 0.7392 \\
\midrule

\textbf{RespLLM (Ours) }                 & \textbf{0.9244} & \textbf{0.9002} & \textbf{0.7958 }& \textbf{0.7840} & \textbf{0.6189} & \textbf{0.6274} & \textbf{1.0000} & \textbf{0.8072} \\

  \bottomrule
  \end{tabular}
\label{tab:ablation:fusion:supervised}
\end{table*}

\begin{table*}[hbtp]
\centering
\footnotesize
\caption{Performance of different fusion methods in our framework when zero-shot testing on test datasets.}
\vspace{3pt}
 \begin{tabular}{lccccc|c|cc}
  \toprule

 Task                   & \bfseries T1    &  \bfseries T2    & \bfseries T3           & \bfseries T4         & \bfseries T5     & \bfseries  Avg.   & \bfseries T6      \\
  \midrule

\textbf{Fusion - Soft}      & 0.6284 & 0.6504 & 0.6550 & 0.6375 & 0.6458 & 0.6434 & - \\
\textbf{Fusion - Add}       & 0.6552 & 0.6396 & 0.5725 & 0.5350 & 0.5500 & 0.5905 & - \\
\textbf{Fusion - CrossAttn} & 0.7272 & 0.7112 & 0.5500 & 0.6125 & 0.5292 & 0.6260 & - \\
\midrule

\textbf{RespLLM (Ours)}         & 0.6424 & 0.6284 & 0.6525 & 0.6750 & 0.6750 & \textbf{0.6547} & 0.5865 \\
  \bottomrule
  \end{tabular}
  \label{tab:ablation:fusion:zeroshot}
\end{table*}

We also compare different open-source LLMs within our framework, with their performance summarized in \cref{tab:ablation:llm:supervised} and \cref{tab:ablation:llm:zeroshot}. The four LLMs show similar AUROCs across tasks, demonstrating the robustness of our training approach. Notably, OpenBioLLM achieves a higher AUROC in the zero-shot setting on the target tasks, likely due to its specialized pre-training on medical corpora, enhancing its diagnostic knowledge for generalized health screening.

\begin{table*}[hbtp]
\centering
\footnotesize
\caption{Performance of different LLMs in our framework when testing on source datasets.}
\begin{tabular}{lcccccccc}
  \toprule
  
 Task                   & \bfseries S1    &  \bfseries S2    & \bfseries S3           & \bfseries S4         & \bfseries S5     & \bfseries S6      & \bfseries S7 & \bfseries Avg.    \\
  \midrule
  \textbf{Gemma2 (2B)}     & 0.9221 & 0.8927 & 0.7555 & 0.7202 & 0.5840 & 0.5709 & 0.9953 & 0.7772          \\
\textbf{Phi-3.5(4B)} & 0.9250 &	0.8989 &	0.7909	& 0.7886	& 0.5964	& 0.6050 &	1.0000 &	0.8007 \\
  \textbf{Mistral (7B)}               & 0.9236 & 0.9006 & 0.7889 & 0.7765 & 0.6040 & 0.6096 & 1.0000 & 0.8005 \\
\textbf{LLaMA (7B)}              & 0.9225      & 0.9055      & 0.7899      & 0.7934      & 0.5986      & 0.6010      & 1.0000       & 0.8016 \\
\textbf{LLaMA3 (8B)}             & 0.9269      & 0.9061      & 0.8048      & 0.7988      & 0.6131      & 0.6171      & 1.0000       & \textbf{0.8095} \\


\midrule
\textbf{OpenBioLLM }                 & 0.9244 & 0.9002 & 0.7958 & 0.7840 & 0.6189 & 0.6274 & 1.0000 & 0.8072 \\

  \bottomrule
  \end{tabular}
  \label{tab:ablation:llm:supervised}
\end{table*}

\begin{table*}[hbtp]
\centering
\footnotesize
\caption{Performance of different LLMs in our framework when zero-shot testing on test datasets.}
\begin{tabular}{lcccccc|cc}
  \toprule

 Task                   & \bfseries T1    &  \bfseries T2    & \bfseries T3           & \bfseries T4         & \bfseries T5     & \bfseries  T6  & \bfseries  Avg.      \\


\midrule

\textbf{Gemma2 (2B)}     & 0.6456 & 0.6256 & 0.6500 & 0.5850 & 0.6833 & 0.5514 & 0.6255 \\
\textbf{Phi (4B)}     & 0.6232 & 0.6200 & 0.5975 & 0.6375 & 0.6583 & 0.5039 & 0.6097 \\
\textbf{Mistral (7B)}    & 0.6264 & 0.6068 & 0.6425 & 0.6575 & 0.6958 & 0.5826 & 0.6368          \\
\textbf{LLaMA (7B)}      & 0.6368 & 0.6340 & 0.6400 & 0.6050 & 0.7083 & 0.5565 & 0.6322 \\
\textbf{LLaMA3 (8B)}     & 0.6388 & 0.6152 & 0.6425 & 0.6625 & 0.6750 & 0.5797 & 0.6372 \\
\midrule
\textbf{OpenBioLLM (8B)} & 0.6424 & 0.6284 & 0.6525 & 0.6750 & 0.6750 & 0.5865 & \textbf{0.6449} \\

  \bottomrule
  \end{tabular}
  \label{tab:ablation:llm:zeroshot}
\end{table*}



\section{Discussion}

In this work, we introduced RespLLM, the first audio-text multimodal LLM for respiratory health prediction. The model not only outperforms state-of-the-art baselines in typical in-distribution testing but also demonstrates stronger generalizability in zero-shot predictions on new datasets and tasks that it was not exposed to during training.

We anticipate that the rise of multimodal LLMs will create exciting opportunities for modality fusion (via Transformers) and for grounding models in heterogeneous data sources (via instruction tuning). Thus, our work serves as a foundational step toward more generalist medical AI models.

\paragraph{Limitations.}
This work presents a proof-of-concept. As such, RespLLM is \textit{not} intended for clinical use and should not be considered safe for such applications. The experiments conducted in this study are limited to respiratory conditions such as COVID-19, COPD, and asthma. We have not tested the model performance on other conditions, such as the flu, due to the limited data available at the moment. However, we hope that such data will become more available in the future, enabling further research.

\paragraph{Future Work} To mitigate the hallucinations that frequently occur in large language models, we replaced the final linear layer in the original LLM with a custom linear layer that only outputs `Yes' or `No' for a given condition. An exciting direction for future work would be to explore the use of the full language model for more comprehensive diagnostics and reasoning in respiratory conditions while maintaining trustworthiness. Additionally, we plan to integrate more biosignal modalities, such as photoplethysmography signals and body temperature dynamics, which could provide a more holistic approach to respiratory health screening.

\bibliographystyle{ACM-Reference-Format}
\pagenumbering{gobble}

\bibliography{bibliography}


\appendix

\section{Data description}\label{apd:first}

\textbf{COVID-19 Sounds~\citep{xia2021covid} }. The COVID-19 Sounds dataset consists of 53,449 audio samples (over 552 hours in total) crowd-sourced from 36,116 participants through the COVID-19 Sounds app. This dataset is comprehensive in terms of demographics and spectrum of health conditions. It also provides participants' self-reported COVID-19 testing status with 2,106 samples tested positive. It consists of three modalities including breathing, cough, and voice recordings. Only breathing and cough modalities are used in this paper.


\textbf{UK COVID-19~\citep{coppock2024audio}}. The UK COVID-19 Vocal Audio Dataset is designed for the training and evaluation of machine learning models that classify SARS-CoV-2 infection status or associated respiratory symptoms using vocal audio. The UK Health Security Agency recruited voluntary participants through the national Test and Trace programme and the REACT-1 survey in England from March 2021 to March 2022, during dominant transmission of the Alpha and Delta SARS-CoV-2 variants and some Omicron variant sublineages. Audio recordings of volitional coughs, exhalations, and speech (speech not included in open access version, nor used in this paper) were collected in the `Speak up to help beat coronavirus' digital survey alongside demographic, self-reported symptom and respiratory condition data, and linked to SARS-CoV-2 test results. 

\textbf{ICBHI~\citep{rocha2019open}}.
The ICBHI Respiratory Sound Database contains audio samples, collected independently by two research teams in two different countries, over several years. Ethical approval was obtained from the ethics committees of the appropriate institutions.

Most of the database consists of audio samples recorded by the School of Health Sciences, University of Aveiro (ESSUA) research team at the Respiratory Research and Rehabilitation Laboratory (Lab3R), ESSUA and at Hospital Infante D. Pedro, Aveiro, Portugal. The second research team, from the Aristotle University of Thessaloniki (AUTH) and the University of Coimbra (UC), acquired respiratory sounds at the Papanikolaou General Hospital, Thessaloniki and at the General Hospital of Imathia (Health Unit of Naousa), Greece.
The database consists of a total of 5.5 hours of recordings in 920 annotated audio samples from 126 subjects.

\textbf{Coswara~\citep{bhattacharya2023coswara}}. The Coswara dataset contains respiratory sounds recorded between April 2020 and February 2022 from 2635 individuals (1819 SARS- CoV-2 negative, 674 positive, and 142 recovered subjects). The respiratory sounds contained nine sound categories associated with variants of breathing, cough and speech. The metadata contains demographic information associated with age, gender and geographic location, as well as the health information relating to the symptoms, pre-existing respiratory ailments, comorbidity and SaRS-CoV-2 test status. 


\textbf{KAUH~\citep{fraiwan2021dataset}}. 
The KAUH dataset includes sounds from seven ailments (i.e., asthma, heart failure, pneumonia, bronchitis, pleural effusion, lung fibrosis, and chronic obstructive pulmonary disease (COPD) as well as normal breathing sounds. The dataset contains the audio recordings from the examination of the chest wall at various vantage points using an electronic stethoscope. The stethoscope placement on the subject was determined by the specialist physician performing the diagnosis. Each recording was replicated three times corresponding to various frequency filters that emphasize certain bodily sounds. The dataset can be used for the development of automated methods that detect pulmonary diseases from lung sounds or identify the correct type of lung sound.


\section{Implementation Details}

\subsection{RespLLM}
\paragraph{Audio encoder.} The audio encoder that we adopt is the pre-trained OPERA-CT model~\citep{zhang2024towards}. It is a  hierarchical token-semantic audio transformer (HTS-AT) model trained with a contrastive learning objective of instance discrimination on respiratory sounds. All audio recordings are padded or cropped to 8 seconds, resampled to 16 kHz and merged into a mono channel. They are then transformed into spectrograms using 64 Mel filter banks with a 64 ms Hann window that shifts every 32 ms, resulting in a spectrogram of $126 \times 64$ dimension. It output patch embeddings of 64 patches, which is input into the LLM as 64 tokens after the alignment module.

\paragraph{LLM and LoRA.} We use the OpenBioLLM model, which has 8B parameters and uses a LLaMA3 architecture. It was developed by Saama AI Lab and released in May 2024 and achieves state-of-the-art performance across various biomedical tasks. To efficiently adapt the LLM model to our tasks, we employ a LoRA module of rank $r = 16$ and $\alpha = 32$.

For the ablation study, we also explored LLaMA-7B~\citep{touvron2023llama}, LLaMA3-8B\footnote{\url{https://ai.meta.com/blog/meta-llama-3/}},
Mistral~\citep{jiang2023mistral}, Gemma-2(2B)\footnote{https://huggingface.co/google/gemma-2-2b} and Phi-3.5\footnote{https://huggingface.co/microsoft/Phi-3.5-mini-instruct}.
\subsection{Baselines}

We use the pre-traiend BERT~\citep{devlin2018bert} for the wording embeddings in the soft fusion baselines, which are of the same dimension of the audio embeddings.







\newpage

\end{document}